\documentclass[mlmain]{jmlr}

\usepackage{booktabs}  
\usepackage{multirow}  
\usepackage{graphicx}

\usepackage{longtable}

\usepackage{booktabs}
\usepackage[load-configurations=version-1]{siunitx} 

\theorembodyfont{\upshape}
\theoremheaderfont{\scshape}
\theorempostheader{:}
\theoremsep{\newline}

\jmlrproceedings{Preprint}{Preprint}
\jmlrvolume{}
\firstpageno{1}
\editors{}

\jmlryear{2025}
\jmlrworkshop{Symmetry and Geometry in Neural Representations}

\title[MindSpace3D]{Dual-Stream EEG Decoding for 3D Visual Perception\titletag{\thanks{Accepted to the Symmetry and Geometry in Neural Representations Workshop (NeurReps) at NeurIPS 2025. To appear in Proceedings of Machine Learning Research (PMLR).}}}

   \author{\Name{Ninon {Lizé Masclef}} \Email{ninon@mit.edu}\\
   \Name{Taisija Demcenko} \Email{taisija@mit.edu}\\
   \Name{Antonella Catanzaro} \Email{antocat@mit.edu}\\
   \Name{Nataliya Kosmyna} \Email{nkosmyna@mit.edu}
\\
   \addr Massachusetts Institute of Technology}

\usepackage{caption} 
\begin{document}

\maketitle

\begin{abstract}

This paper explores a novel brain decoding model for 3D shape perception through a dual pathway architecture mirroring biological vision. Our bio-inspired approach implements separate decoding modules for object identity and spatial orientation, inspired by ventral and dorsal pathways, during continuous rotations.
We employ circular regression for angle prediction and develop EEG-conditioned multiview diffusion for 3D reconstruction. Our approach successfully decodes both object identity and spatial orientation from EEG signals and enables 3D reconstruction from neural activity, with interpretability analyses revealing temporally structured involvement of ventral, dorsal, and motor‑related channels rather than a static ventral dominance in supporting object and angle decoding.


\end{abstract}

\section{Introduction}
\label{sec:intro}

Neural networks for visual perception have drawn fundamental inspiration from biological vision systems, particularly their hierarchical organization \citep{hubel_receptive_1962, fukushima_neocognitron_1980, lehky_network_1988}. While convolutional neural networks show correspondence with the ventral visual pathway --- the brain's ``what'' system for object recognition \citep{dicarlo_how_2012, cichy_comparison_2016} --- they struggle with 3D spatial and geometric understanding \citep{hinton_matrix_2018, bronstein_geometric_2017}. This limitation is particularly evident in 3D tasks: standard CNNs are biased toward texture over shape \citep{geirhos_imagenet-trained_2018}, and humans significantly outperform CNNs in cross-viewpoint 3D shape matching \citep{oconnell_approaching_2023}. A similar challenge arises in brain decoding, where models struggle to differentiate viewpoints within object classes and extract orientation features from EEG signals \citep{li_visual_2024}.

We address this limitation with a dual‑stream decoding architecture inspired by ventral and dorsal visual pathways, implementing separate modules for object identity and spatial orientation. Our approach reconstructs rotating 3D objects from EEG by explicitly modeling viewpoint‑tolerant object recognition and viewpoint‑dependent spatial transformations, decoding object identity and angular positioning to provide novel insight into geometric representation in biological and artificial systems. Our results demonstrate that EEG signals preserve both object identity (up to 68\% accuracy across six semantic categories) and spatial orientation (10-11° mean absolute error) during 3D perception, with interpretability analysis revealing time‑resolved integration across ventral, dorsal, and motor‑related pathways, consistent with evidence for dynamic dorsal–ventral interactions and visuo‑motor coupling in 3D object perception.


\section{Related Works}
\label{sec:related}
\subsection{3D Shape Perception}
Human 3D object perception involves hierarchical processing in specialized areas, beginning in early visual areas (V1-V3) which extract basic 2D features such as edges, orientation, and contrast \citep{dicarlo_how_2012}. 3D perception differs from 2D perception by its complexity integrating spatial, depth and structural information, engaging stereo cues such as binocular disparity, leveraging the slight differences in images received by each eye to provide robust depth information \citep{julesz_foundations_1971}. Current evidence indicates this information then diverges into two distinct but interconnected pathways: the ventral stream (``what'' pathway) projects from V1 through V2 and V4 to the inferior temporal cortex, and is crucial for object recognition, form representation, and high-level semantic processing \citep{mishkin_object_1983}. In parallel, the dorsal stream (initially ``where'', later ``how'' pathway \citep{goodale_separate_1992}) projects from V1 through V2 and regions like MT/V5 to the posterior parietal cortex. This pathway specializes in spatial relationships, motion, and action-relevant information, and critically transforms visual input into egocentric coordinate systems necessary for motor planning and execution \citep{gallivan_neural_2015, goodale_evolving_2004}.
Despite their distinct specializations, dorsal and ventral streams interact extensively in attention \citep{chica_neural_2013} and visual mental imagery \citep{spagna_visual_2023}. Neurons in dorsal regions can respond selectively to 3D contours and surfaces \citep{theys_shape_2015}, and neuroimaging and lesion work shows substantial cross‑talk, with object and spatial information represented in both streams \citep{kravitz_new_2011, vaziri-pashkam_goal-directed_2017, bartolomeo_chapter_2022}. This motivates models that treat ventral-dorsal processing as interacting components rather than a strict dichotomy.


\subsection{Bio-inspired 3D Brain Decoding}

Lehky and Sejnowski (1988) demonstrated that neural networks trained to extract 3D shape from shading develop cortex-like receptive fields, suggesting that the computational demands of 3D vision may drive similar representational solutions across artificial and biological neural networks. This observation has motivated extensive research into bio-inspired architectures for visual processing \citep{lehky_network_1988}. Convolutional neural networks (CNNs) trained on large naturalistic image collections show correspondence with primate ventral visual stream activity patterns \citep{guclu_deep_2015, yamins_performance-optimized_2014}, with deeper layers responding to increasingly complex features mirroring the V1-to-IT progression. Performance-optimized hierarchical models accurately predict neural responses in IT and V4 without explicit neural constraints, suggesting that categorization objectives alone may drive brain-like representations \citep{khaligh-razavi_deep_2014}.
However, divergences emerge when examining transformation tolerance mechanisms. While CNNs match human accuracy on object categorization, they differ from biological systems in how they handle spatial transformations \citep{xu_understanding_2022}. Human visual areas show progressively more transformation‑tolerant and consistent representations from V1 to higher regions, whereas CNNs often lose representational consistency at deeper layers, suggesting a reliance on ‘brute‑force’ memorization rather than geometric structure. This algorithmic divergence matters for brain decoding: human‑aligned encoders that better match cortical representational geometry can improve EEG/MEG decoding performance \citep{rajabi_human-aligned_2025}.

Effective bio-inspired architectures require explicit integration of geometric and perceptual principles, particularly for dorsal-ventral pathway modeling where spatial processing and transformation tolerance are computational requirements. Neural networks for 3D processing span several paradigms: volumetric CNNs \citep{maturana_voxnet_2015}, geometric deep learning on non‑Euclidean domains \citep{bronstein_geometric_2017}, attention‑based point‑cloud models such as Point Transformer \citep{zhao_point_2021}, and implicit neural representations like NeRF, Occupancy Networks, and DeepSDF \citep{mildenhall_nerf_2020, mescheder_occupancy_2019, park_deepsdf_2019}. Recent approaches incorporate explicit geometric symmetries, for example capsules \citep{sabour_dynamic_nodate, hinton_matrix_2018}, SO(3)-equivariant networks \citep{thomas_tensor_2018}, and steerable CNNs \citep{cohen_group_2016, weiler_3d_nodate}. Understanding how such inductive biases translate to brain decoding requires examining the geometric structure preserved in neural signals.
\subsection{3D Geometric Inductive Bias}
Human 3D shape perception demonstrates the brain's intrinsic geometric processing capabilities through mechanisms that achieve multiview consistency while revealing systematic inductive biases. The visual system's foundation lies in its topographic organization, e.g. retinotopic maps in early visual areas (V1-V3) maintain precise spatial correspondence between functional and cytoarchitectonic borders, creating an inherent geometric framework for processing orientation, disparity, and spatial features \citep{tsao_patches_2008, tsutsui_integration_2001}. This spatial organization enables hierarchical geometric processing and depth cue combination, where low-level shape features are decoded within 60~ms \citep{foxe_flow_2002}, building toward complex 3D representations through integration of binocular disparity, perspective, shading, and motion information in higher areas like LOC and hMT+/V5 \citep{lehky_network_1988, beeck_perceived_2008}.
This processing is so fundamental that we automatically perceive 2D sketches as 3D objects, unable to suppress our geometric interpretation even when consciously aware of the flat medium \citep{torralba_foundations_2024}.

However, human 3D perception is not perfectly viewpoint-invariant, as object recognition speed correlates with rotation angle from canonical poses
\citep{shepard_mental_1971, palmer_canonical_1981}, revealing innate inductive biases encoded in domain-specific connectivity patterns that facilitate rapid object recognition while predisposing the brain to extract stable 3D representations from ambiguous inputs \citep{welchman_3d_2005}. These biases reflect equivariant neural representations where transformations in visual input produce corresponding transformations in neural activity patterns, enabling consistent object recognition while maintaining sensitivity to spatial relationships crucial for action and navigation.

These intrinsic geometric processing capabilities are reflected in EEG signals, which capture the synchronized activity of geometrically aligned pyramidal cells across visual cortex. Despite EEG's limited spatial resolution for detailed reconstruction \citep{halliday_changes_1970}, its high temporal resolution enables decoding of visual features through visual-evoked potential components, such as N1, P2, and N2 components, specifically involved in processing 3D shape and depth, showing hemispheric dissociation during the N1/N2 complex for 3D shape (left hemisphere) and depth (right hemisphere) \citep{kasai_event-related_2001}. These components also provide insights into low-level visual features including visual field position \citep{halliday_changes_1970, jeffreys_source_1972}.
The geometric sensitivity of EEG extends to complex spatial cognition: \cite{kashihara_evaluation_2011}    
 demonstrated that EEG effectively captures neural signatures during both 2D and 3D mental imagery tasks, suggesting preserved information for distinguishing geometrical shapes across dimension. This spatial sensitivity follows the brain's hierarchical organization: ablation experiments in visual brain decoding confirmed that occipital regions achieve highest accuracy and reconstruction performance compared to the temporal, parietal and frontal areas, directly reflecting the visual processing hierarchy \citep{li_visual_2024}.
 
 Recent advances demonstrate the feasibility of geometric decoding from brainwave activity: primitive 3D shapes \citep{esfahani_classification_2012}, colored geometric symbols \citep{bang_spatio-spectral_2021}, EEG-based 3D object reconstruction as colored point cloud \citep{guo_neuro-3d_2024}, 3D reconstruction from fMRI \citep{gao_mind-3d_2024}, and EEG-driven 3D object generation using latent diffusion models \citep{xiang_eeg-driven_2024}. However, current EEG-based 3D decoding approaches treat perception as a single pathway problem, failing to leverage the geometric structure underlying biological vision. From a geometric perspective, 3D shape understanding requires disentangling invariant object properties from equivariant spatial transformations; this decomposition is embodied in the distinct but interacting functional roles of the ventral ``what'' and dorsal ``where/how'' pathways. We develop a dual-stream architecture that explicitly models this geometric decomposition, enabling principled 3D reconstruction from EEG through separate object identity classification and angular transformation regression.

\section{Methods}
\label{sec:method}

To validate this bio‑inspired architecture, we decompose 3D visual decoding into three complementary tasks that aim to mirror, as closely as possible, proposed biological processing stages: object classification (ventral stream), spatial orientation regression (dorsal stream), and their integration for 3D reconstruction.

\subsection{Dataset and Experimental Design}

Eleven (11) participants were recruited for this study. Neural activity was recorded using a 64-channel ANT Neuro wet electrode system positioned according to the international 10-20 system at 512~Hz sampling frequency, with impedances maintained below 20~k$\Omega$. With 11 subjects, the main group‑level effect we report corresponds to a very large effect size (Cohen’s 
$d \approx 2.8$), indicating that the study is highly powered at $\alpha = 0.05$. Participants viewed rotating 3D object stimuli in an immersive virtual reality environment rendered using Unity and presented via Meta Quest 2 head-mounted display. The VR paradigm was specifically chosen to leverage binocular disparity cues and enable natural depth perception mechanisms unavailable in conventional 2D presentation modalities. The stimulus set comprised 78 distinct 3D objects uniformly distributed across six semantic categories (13 examples per category): organic natural objects (banana, strawberry), manufactured objects (basketball), and biological entities (human face, panda, tiger). Each object underwent continuous 360° rotation over an 8-second presentation window with angular velocity $\frac{d\theta}{dt} = \frac{\pi}{4}$ rad/s, followed by a 2~s inter-stimulus interval to mitigate potential neural adaptation effects. Example of stimuli is displayed in \figureref{fig:1ab}. To control for temporal confounds and sequence-dependent neural responses, stimulus presentation order was fully randomized across participants and recording sessions. Each participant completed one $\sim$65-minute session with 78 objects $\times$ 6 repetitions (468 trials per subject, 5{,}148 trials total). 
This experimental design enables us to separately analyze object identity across rotations and spatial transformations from EEG, providing well‑defined labels for both geometric invariance (object identity) and equivariance (rotation angle).

\subsection{Model Architectures and Training}

We benchmarked three state-of-the-art architectures: EEGNet \citep{lawhern_eegnet_2018}, a compact CNN for visual evoked potential decoding; EEGConformer \citep{song_eeg_2023}, a hybrid CNN-transformer for long-range temporal dependencies; and CTNet \citep{zhao_ctnet_2024}, a convolutional transformer combining spatial feature extraction with multi-head attention. These architectures offer complementary strengths for processing spatiotemporal EEG features across extended time windows. Hyperparameters were optimized per subject using Bayesian optimization \citep{snoek_practical_2012}. For EEGNet, we increased the depthwise kernel length to 256 samples to better capture visual event-related potentials \citep{luck_introduction_2014}. Model performance was assessed via mean accuracy across 5-fold cross-validation.

\subsubsection{Circular Regression Architecture}

To enable rotation angle estimation, EEGNet was adapted for circular regression by replacing the classification head with a 2-dimensional output layer predicting unit vectors $[\sin(\theta), \cos(\theta)]$. This geometric representation naturally handles periodic boundary conditions and avoids discontinuities near $\pm\pi$, ensuring angles differing by $2\pi$ are treated as identical. The circular regression loss computes squared chord length between predicted and target unit vectors:
\begin{equation}
L_{\text{circular}} = \frac{1}{B}\sum_{i=1}^{B} \left[1 - \cos(\hat{\theta}_i - \theta_i)\right]
\end{equation}
where $B$ is batch size, equivalent to mean squared error in circular space. Classification and angle‑regression models were instantiated as separate EEGNet networks with no weight sharing across tasks.

\subsubsection{EEG-Conditioned 3D Reconstruction}

\begin{center}
\centering
\begin{figure}[h!]
\includegraphics[width=1.\linewidth]{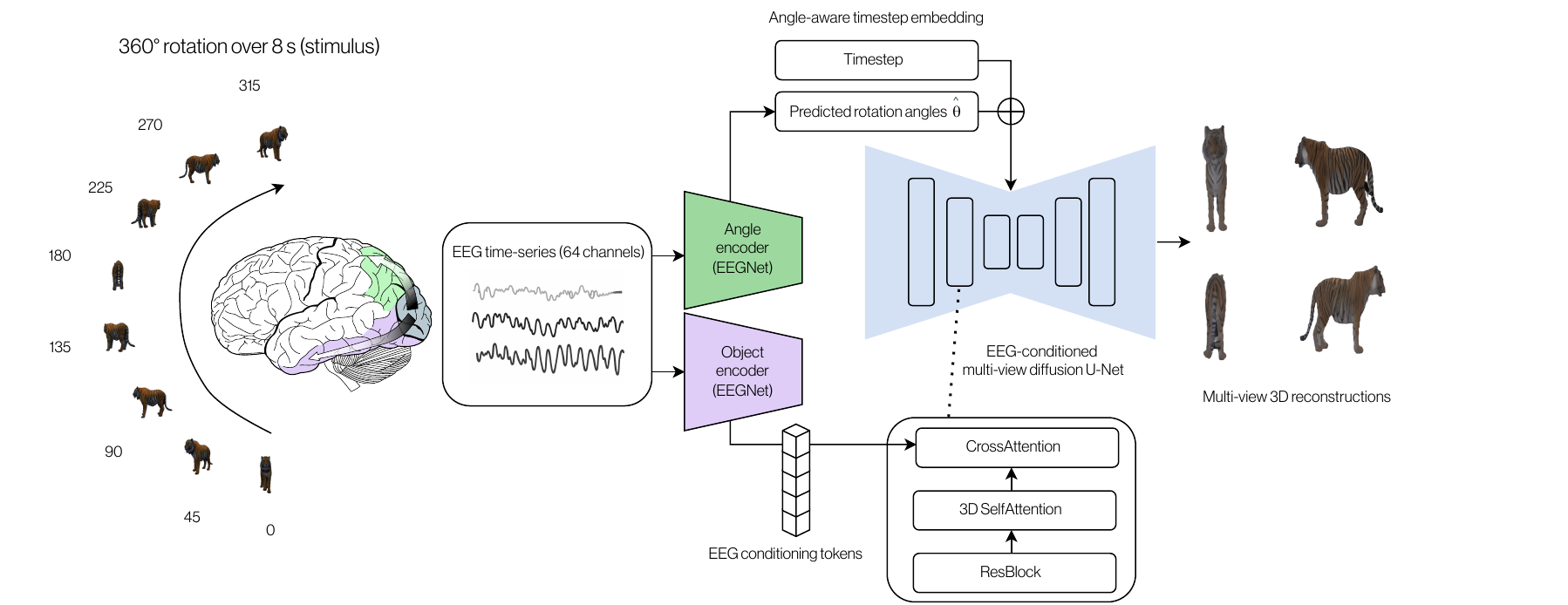}\hfill
\caption{Proposed dual‑stream architecture for 3D brain decoding from EEG. \vspace{-.5cm}
}\label{fig:1ab}
\end{figure}    
\end{center}
\vspace{-1cm}
We developed a dual-stream 3D reconstruction pipeline that integrates object classification and angle regression to condition a multiview diffusion model, mirroring biological integration of object identity (``what'') and spatial orientation (``where'') pathways for 3D shape generation from neural signals. We adapt MVDream \citep{shi_mvdream_2024}, a text-to-multiview diffusion model that employs an inflated 3D self‑attention mechanism to extend pretrained 2D self‑attention into multi‑view 3D attention, promoting consistent generation across multiple camera viewpoints. A pretrained, frozen EEGNet object classifier is repurposed as a feature extractor: its classification head is replaced by a 512‑dimensional embedding layer whose output is projected to 1024 dimensions via a fixed random linear layer matching CLIP’s text‑embedding space. In the diffusion U‑Net, multi‑view latents provide the queries for cross‑attention, while the projected EEG embedding supplies the keys and values, enabling EEG‑conditioned 3D‑aware denoising. An angle encoder predicts EEG‑derived rotation angles that are injected through an angle‑aware timestep embedding. The diffusion model is fine‑tuned for EEG‑conditional generation using Low‑Rank Adaptation (LoRA \citep{hu_lora_2021}; $r=24$ $\alpha=48$) applied to attention and feed‑forward layers, preserving the pretrained multi‑view prior while enabling efficient adaptation. The multi‑view diffusion loss is:


\begin{equation}
\mathcal{L}_{\text{MV}}(\theta) = \mathbb{E}_{t,\epsilon} \left[ \|\epsilon - \epsilon_\theta(x_t; f_{\text{obj}}, \hat{\theta}, t)\|_2^2 \right]
\end{equation}

where $x_t$ are noisy multi‑view latents at diffusion step $t$, $f_{\text{obj}}$ denotes the EEG‑derived object identity embedding used as conditioning tokens, and $\hat{\theta}$ are the predicted rotation angles injected through an angle‑aware timestep embedding.
We fine‑tune for up to 2{,}000 epochs on a single A100 GPU with early stopping based on validation loss, using batch size $32$ and learning rate 
$10^{-4}$; generation uses 50 denoising steps with the same scheduler as during training.




\vspace{-2mm}

\subsection{Channel Activation Analysis}

To investigate whether EEGNet captures the functional specialization of dorsal (spatial processing) and ventral (object identity) visual pathways in 3D shape perception, we applied Gradient-weighted Class Activation Mapping \citep{selvaraju_grad-cam_2020} to the convolutional layers. GradCAM computes spatiotemporal importance maps as:

\begin{equation}
\text{GradCAM}(i,j) = \text{ReLU}\left(\sum_{k} \alpha_k \cdot A_{k,i,j}\right)
\end{equation}

where $\alpha_k = \frac{1}{HW} \sum_{m,n} \frac{\partial y}{\partial A_{k,m,n}}$ represents the average gradient of the target output $y$ (object class or rotation angle $\hat{\theta}$) with respect to feature map $k$, and $A_{k,i,j}$ denotes activations at spatial position $(i,j)$. 

To assess regional contributions, we defined three anatomically-motivated channel subsets of equal size (10 channels each): \textit{dorsal} pathway (Cz, C1, C2, Pz, P1, P2, CP1, CP2, POz, Oz), \textit{ventral} pathway (T7, T8, TP7, TP8, P7, P8, PO7, PO8, O1, O2), and \textit{motor/premotor} regions (C3, C4, C5, C6, FC3, FC4, FC1, FC2, CP5, CP6). Channels were ranked by total GradCAM activation, and top bias scores were calculated as:

\begin{equation}
\text{Top Bias} = \frac{1}{|C|}\sum_{i\in C}{1_{p_i < N/2}}
\end{equation}

where $C$ represents the channel subset, $p_i$ is channel $i$'s rank, and $N=64$ total channels. We hypothesized that object classification would engage ventral-like patterns, while angle regression would reveal dorsal-like signatures, and explored whether these contributions would vary over time across ventral, dorsal, and motor-related subsets.

\section{Results}
\label{sec:results}

{\bfseries 3D Shape Classification:} We evaluated whether EEG signals preserve object identity despite continuous spatial transformations by classifying rotating 3D objects into six semantic categories from 8‑second epochs. All three architectures achieved above‑chance performance (chance level: 16.7\%), with EEGNet yielding the highest mean accuracy, followed by CTNet and EEGConformer (Table~\ref{tab:modelacc}). Across subject‑specific models, performance varied and some models showed instability in cross‑validation (e.g., Sub5: SD = 0.03; Sub6: SD = 0.10), but a consistent ranking emerged: Sub5 achieved the highest accuracy, followed by Sub6 and Sub11 (for EEGConformer, Sub9 was third). The EEGNet model trained for Sub5 was used for subsequent 3D reconstruction, as it combined the best validation and test accuracy with the most stable cross‑validation performance.


\begin{table}[hbtp]
\centering
  {\caption{
  3D shape classification and angle regression performance from EEG signals (mean ± SD across 5‑fold CV).
  }\label{tab:modelacc}} \vspace{4mm} 
\begin{tabular}{l|ccc|c}
\toprule
\bfseries Subject & \bfseries EEGNet (acc) & \bfseries EEGConformer (acc) & \bfseries CTNet (acc) & \bfseries Angle (MAE) \\
\midrule
Sub1  & 0.31 $\pm$ 0.05 & 0.30 $\pm$ 0.03 & 0.27 $\pm$ 0.03 & 11.1 $\pm$ 0.03 \\
Sub2  & 0.36 $\pm$ 0.09 & 0.19 $\pm$ 0.02 & 0.40 $\pm$ 0.04 & 11.0 $\pm$ 0.08 \\
Sub3  & 0.29 $\pm$ 0.03 & 0.29 $\pm$ 0.03 & 0.30 $\pm$ 0.05 & 10.8 $\pm$ 0.04 \\
Sub4  & 0.30 $\pm$ 0.05 & 0.34 $\pm$ 0.04 & 0.26 $\pm$ 0.03 & 10.9 $\pm$ 0.09 \\
Sub5  & 0.68 $\pm$ 0.03 & 0.56 $\pm$ 0.05 & 0.56 $\pm$ 0.03 & 10.7 $\pm$ 0.04 \\
Sub6  & 0.46 $\pm$ 0.10 & 0.54 $\pm$ 0.07 & 0.54 $\pm$ 0.06 & 10.8 $\pm$ 0.06 \\
Sub7  & 0.32 $\pm$ 0.06 & 0.29 $\pm$ 0.05 & 0.30 $\pm$ 0.04 & 11.1 $\pm$ 0.12 \\
Sub8  & 0.32 $\pm$ 0.05 & 0.22 $\pm$ 0.02 & 0.39 $\pm$ 0.07 & 10.9 $\pm$ 0.12 \\
Sub9  & 0.37 $\pm$ 0.08 & 0.39 $\pm$ 0.03 & 0.34 $\pm$ 0.04 & 11.1 $\pm$ 0.08 \\
Sub10 & 0.34 $\pm$ 0.08 & 0.37 $\pm$ 0.08 & 0.39 $\pm$ 0.04 & 10.9 $\pm$ 0.06 \\
Sub11 & 0.43 $\pm$ 0.07 & 0.32 $\pm$ 0.05 & 0.42 $\pm$ 0.08 & 11.1 $\pm$ 0.04 \\
\bottomrule
\end{tabular}
\end{table}
\noindent{\bfseries Rotation Angle Regression:} We evaluated the dorsal stream component by regressing rotation angles from EEG windows of varying durations ($T_w \in \{512\}$ samples, corresponding to 1s epochs). The target angle was computed as $\theta_{\text{true}} = \frac{\pi}{4} \times t_{\text{window\_center}}$ where $t_{\text{window\_center}}$ represents the temporal center of each analysis window. In contrast to object classification, all angle regression models exhibited comparable error rates, with a 1~s window size proving most effective for learning angular information. This finding is consistent with the expectation that larger temporal windows encompass a greater range of angular positions, thereby reducing prediction accuracy for specific angles. Within the 1~s window, the angular distance spans $45^\circ$. The achieved MAE values ($10-11^\circ$) across all models demonstrate robust performance in angular prediction.

\vspace{1mm}\noindent{\bfseries Interpretability Analysis:} Subject-level top‑bias scores (\figureref{fig:topbias}) show that the top four object decoders exhibit equivalent or stronger ventral and motor activation compared to dorsal channels, although this pattern is heterogeneous across participants. Paired t‑tests ($\alpha$ = 0.05) confirmed that ventral, dorsal, and motor subsets contribute comparably at the group level across 1s windows, indicating that the observed motor dominance is not a trivial artifact of isolated channel noise but reflects a systematic pattern across subjects. Grad‑CAM therefore indicates that decoding performance is not driven by a static binary ‘ventral‑only’ versus ‘dorsal‑only’ readout but by how models distribute attention across dorsal, ventral, and motor channels over time; mirroring human dorsal-ventral integration dynamics \citep{ayzenberg_temporal_2023}. 
\begin{figure}[h!]
\floatconts
  {fig:topbias}
  {\vspace{-0.5cm}
  \caption{Top bias scores of different channel subsets, with subjects sorted from best performing model (left) to worse performing (right).}}
  {\includegraphics[width=0.8\textwidth]{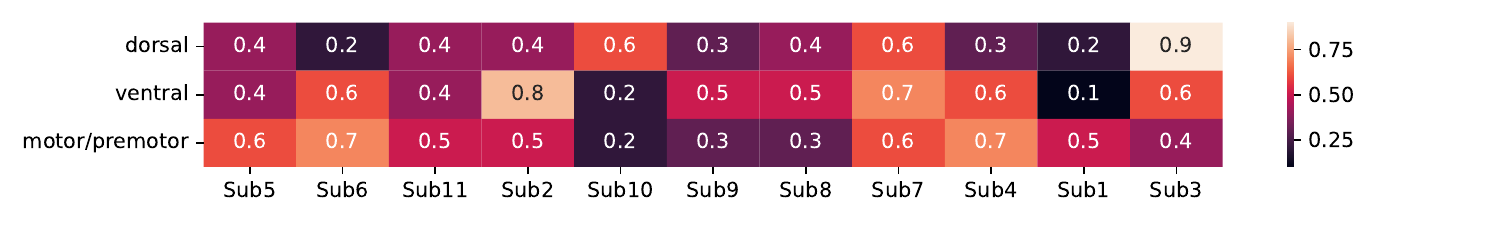}}
\end{figure}
Specifically, the dissociation between object and angle models suggests a partial disentangling of invariant and equivariant components: for angle decoders, accuracy benefits from early, positive motor bias together with a relaxing dorsal penalty, as dorsal correlations move from negative values toward approximately zero ($p<0.05$); whereas object decoders are penalized when they overweight ventral channels early and dorsal channels late, consistent with more invariant object information arising from a balanced, temporally structured integration across dorsal, ventral, and motor‑related subspaces rather than from a single ventral‑like code \citep{farivar_dorsalventral_2009}. 
Beyond subject‑specific top‑bias patterns, Grad‑CAM maps show substantial between‑subject variability (Appendix~\ref{apd:first}), with a small subset of occipital–parietal and frontal channels displaying both high mean activation and high variance, while several midline posterior sites remain consistently low‑importance. This suggests that dual‑stream decoding relies on overlapping but individually idiosyncratic sensor configurations.


\vspace{1mm}\noindent{\bfseries 3D Reconstruction:} Table~\ref{tab:rec} presents results on the held‑out test set. Our method achieves SSIM of 0.856±0.038 and LPIPS of 0.275±0.061, slightly exceeding validation performance (SSIM: 0.833±0.035, LPIPS: 0.297±0.058) and indicating good generalization. 
Performance remains consistent across object categories (SSIM: 0.840--0.876; Appendix~\ref{tab:per_class_appendix}) and viewpoints (SSIM: 0.841--0.879; Appendix~\ref{tab:per_view_full}), suggesting that the EEG‑conditioned multiview prior captures a viewpoint‑dependent but geometrically coherent shape representation. Together with the channel‑level Grad‑CAM analysis, this supports a picture in which ventral, dorsal, and motor‑related channels jointly provide the invariant (identity) and equivariant (angle) information required for consistent 3D reconstruction. Figure~\ref{fig:rec} shows representative reconstructions, where generated objects preserve class‑specific features (facial structure, organic forms) while maintaining geometric consistency across viewpoints, demonstrating the feasibility of EEG‑conditioned neural‑to‑3D reconstruction from non‑invasive signals.

\begin{table}[hbtp]
\centering
\caption{Quantitative evaluation of EEG-to-multiview reconstruction}
\vspace{4mm} 
\begin{tabular}{lccc} 
\toprule
\textbf{Split} & \textbf{PSNR} $\uparrow$ & \textbf{SSIM} $\uparrow$ & \textbf{LPIPS} $\downarrow$ \\
\midrule
Validation & $15.82 \pm 0.95$ & $0.833 \pm 0.035$ & $0.297 \pm 0.058$ \\
Test & $\mathbf{16.30 \pm 0.73}$ & $\mathbf{0.856 \pm 0.038}$ & $\mathbf{0.275 \pm 0.061}$ \\
\bottomrule
\end{tabular}
\label{tab:rec} 
\end{table}

\begin{center}

\vspace{-.7cm}
\begin{figure}[h!]
\centering 
\includegraphics[width=\linewidth]%
{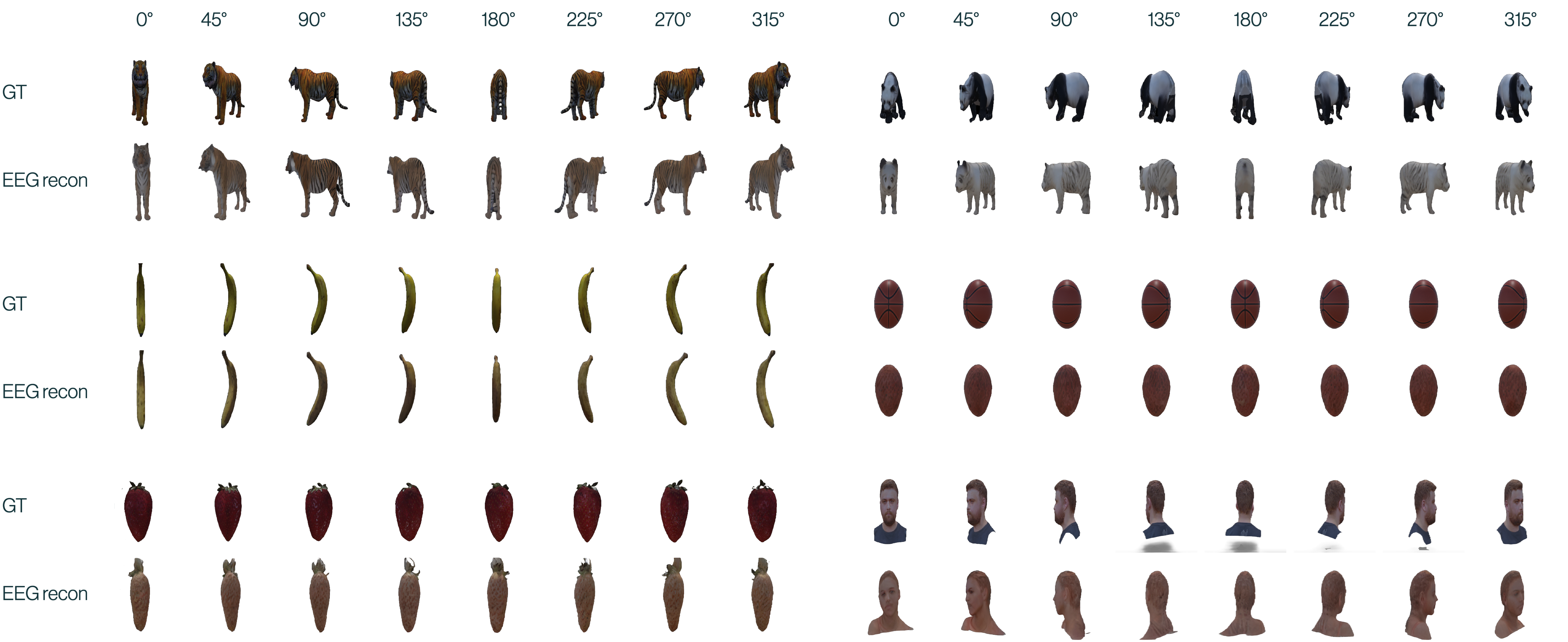}\hfill
\caption{Multi-view test set reconstructions across 6 categories 
(ground truth: top, generated: bottom). Each shows 8 viewpoints 
(0°--315°). 
}\label{fig:rec}
\end{figure}    
\end{center}

\vspace{-1.4cm}

\section{Conclusion}
\label{sec:ccl}

This work makes three primary contributions to neural decoding and bio-inspired 3D vision: (1) We demonstrate the first dual-stream architecture for 3D brain decoding that uses separate modules for object identity (viewpoint‑tolerant ``what'') and spatial transformation (viewpoint‑dependent ``where/how'') processing, achieving robust performance on both classification (up to 68\% accuracy) and angular regression (10-11° MAE); (2) 
We provide interpretable evidence for dynamic dorsal–ventral–motor involvement in EEG through Grad-CAM, showing that successful decoders avoid simple ventral dominance and instead rely on early motor engagement with time‑varying dorsal-ventral contributions; and 
(3) We establish the feasibility of EEG-conditioned 3D reconstruction through multiview diffusion, enabling direct generation of 3D objects from neural signals.
Further investigation of geometric priors through comparison with steerable CNNs and equivariant architectures would quantify the benefits of biological versus purely geometric inductive biases for 3D brain decoding. Quantitative evaluation through single-view versus multiview reconstruction metrics and analysis of rotational equivariance properties would further validate the geometric understanding captured by our dual-stream approach. 


\begin{acks} \vspace{-.2cm}
We thank Constanze Albrecht, Stephanie Chen, and Emmie Fitz-Gibbon for their assistance with data collection. Ninon was supported by an O'Shaughnessy Fellowship. We are grateful to Manuel Cherep and to the reviewers for feedback that improved this manuscript.
\end{acks}

\bibliography{references}

\clearpage
\appendix

\section{Gradient-weighted Class Activation Mapping 1-second windows for best 3D object recognition model per subject}\label{apd:first}

\begin{figure}[htbp]
\floatconts
  {fig:gradcamheads}
  {\vspace{-0.5cm}
  }
  {\includegraphics[width=0.9\textwidth]{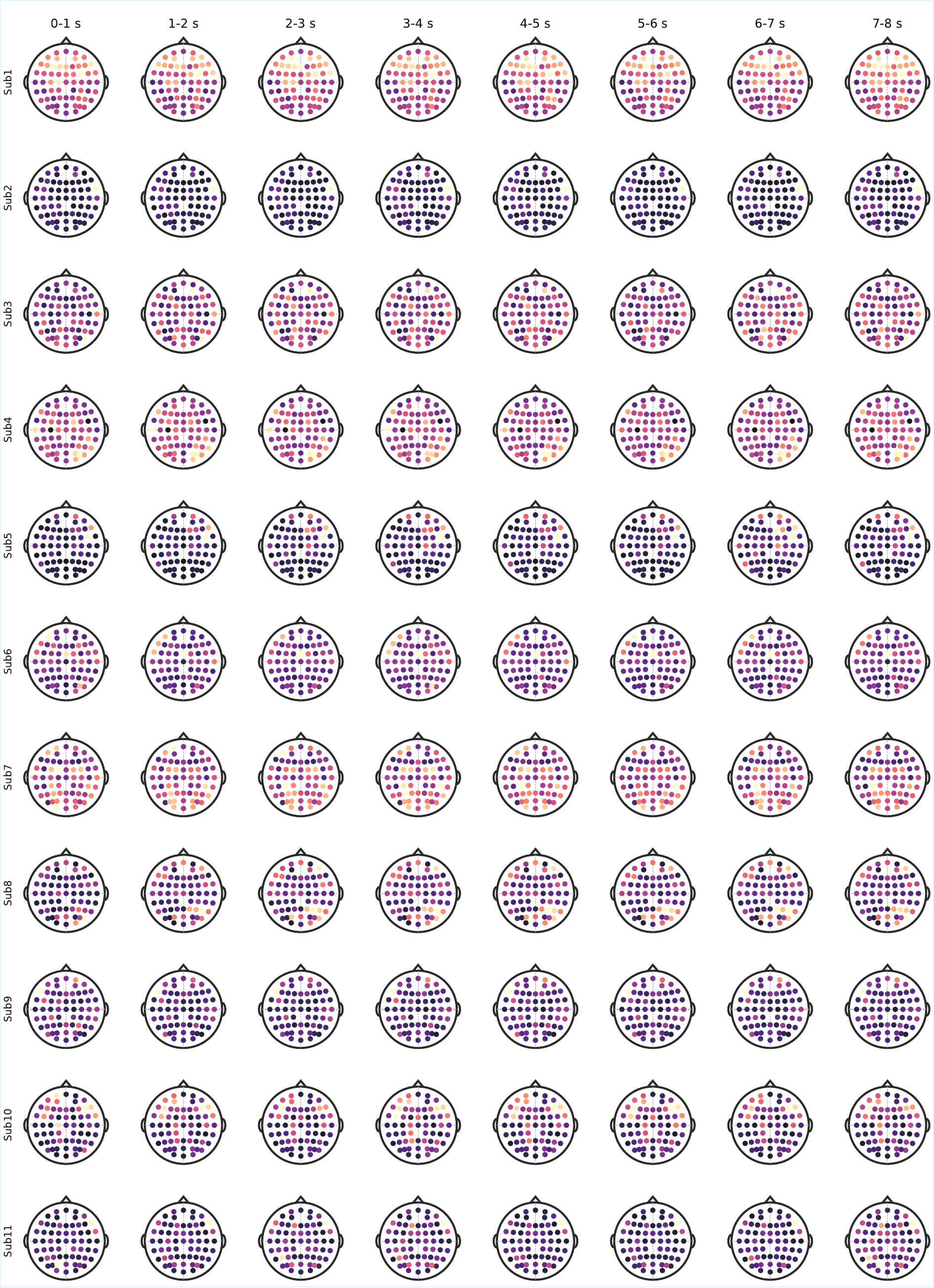}}
\end{figure}
\bigbreak
\section{Per-class reconstruction performance}\label{apd:second}
\begin{center}
\begin{table}[h]
\centering
\caption{Per-class performance on test set. 
}
\vspace{5mm} 
\label{tab:per_class_appendix}
\begin{tabular}{lccc}
\toprule
\textbf{Class} & \textbf{PSNR} $\uparrow$ & \textbf{SSIM} $\uparrow$ & \textbf{LPIPS} $\downarrow$ \\
\midrule
Banana        & $16.50 \pm 0.77$ & $0.868 \pm 0.041$ & $0.249 \pm 0.049$ \\
Basketball    & $15.77 \pm 0.68$ & $0.845 \pm 0.028$ & $0.292 \pm 0.040$ \\
Face          & $16.38 \pm 0.80$ & $0.876 \pm 0.027$ & $0.282 \pm 0.078$ \\
Panda         & $16.29 \pm 0.99$ & $0.868 \pm 0.031$ & $0.233 \pm 0.040$ \\
Strawberry    & $16.01 \pm 0.60$ & $0.840 \pm 0.030$ & $0.306 \pm 0.058$ \\
Tiger         & $16.52 \pm 0.41$ & $0.851 \pm 0.041$ & $0.280 \pm 0.064$ \\
\midrule
\textbf{Overall} & $16.30 \pm 0.73$ & $0.856 \pm 0.038$ & $0.275 \pm 0.061$ \\
\bottomrule
\end{tabular}
\end{table}
\end{center}
\section{Per-View Consistency}\label{apd:third}

\begin{center}

\begin{table}[h]  
\centering
\caption{Complete per-view metrics on validation and test sets.}
\label{tab:per_view_full}
\vspace{5mm} 
\small

\resizebox{\textwidth}{!}{%
\begin{tabular}{clcccccc}
\toprule
\multirow{2}{*}{\textbf{View}} & \multirow{2}{*}{\textbf{Angle}} & \multicolumn{3}{c}{\textbf{Validation}} & \multicolumn{3}{c}{\textbf{Test}} \\
\cmidrule(lr){3-5} \cmidrule(lr){6-8}
& & PSNR $\uparrow$ & SSIM $\uparrow$ & LPIPS $\downarrow$ & PSNR $\uparrow$ & SSIM $\uparrow$ & LPIPS $\downarrow$ \\
\midrule
0 & 0°   & $16.12 \pm 1.19$ & $0.862 \pm 0.031$ & $0.261 \pm 0.063$ & $16.46 \pm 0.84$ & $0.875 \pm 0.026$ & $0.251 \pm 0.066$ \\
1 & 45°  & $15.84 \pm 0.95$ & $0.828 \pm 0.040$ & $0.303 \pm 0.058$ & $16.27 \pm 0.78$ & $0.852 \pm 0.044$ & $0.282 \pm 0.065$ \\
2 & 90°  & $15.56 \pm 0.96$ & $0.811 \pm 0.048$ & $0.323 \pm 0.062$ & $16.12 \pm 0.77$ & $0.841 \pm 0.053$ & $0.295 \pm 0.069$ \\
3 & 135° & $15.71 \pm 1.01$ & $0.826 \pm 0.042$ & $0.299 \pm 0.058$ & $16.33 \pm 0.70$ & $0.853 \pm 0.042$ & $0.276 \pm 0.060$ \\
4 & 180° & $16.12 \pm 1.19$ & $0.866 \pm 0.034$ & $0.259 \pm 0.070$ & $16.55 \pm 0.81$ & $0.879 \pm 0.027$ & $0.244 \pm 0.064$ \\
5 & 225° & $15.71 \pm 1.00$ & $0.826 \pm 0.040$ & $0.307 \pm 0.064$ & $16.23 \pm 0.76$ & $0.852 \pm 0.044$ & $0.281 \pm 0.063$ \\
6 & 270° & $15.66 \pm 0.89$ & $0.816 \pm 0.044$ & $0.322 \pm 0.062$ & $16.13 \pm 0.76$ & $0.843 \pm 0.049$ & $0.292 \pm 0.068$ \\
7 & 315° & $15.79 \pm 1.04$ & $0.828 \pm 0.039$ & $0.306 \pm 0.058$ & $16.27 \pm 0.75$ & $0.853 \pm 0.040$ & $0.280 \pm 0.063$ \\
\midrule
\multicolumn{2}{l}{\textbf{Mean}} & $15.82 \pm 1.03$ & $0.833 \pm 0.040$ & $0.297 \pm 0.061$ & $16.30 \pm 0.77$ & $0.856 \pm 0.040$ & $0.275 \pm 0.063$ \\
\bottomrule
\end{tabular}
}
\end{table}

\end{center}

\end{document}